\documentclass[runningheads]{llncs}
\RequirePackage{silence}  
\usepackage{graphicx}
\usepackage{comment}
\usepackage{amsmath,amssymb}
\usepackage{color}
\usepackage{url}
\usepackage{hyperref}

\usepackage{multirow}
\usepackage{tikz}
\usepackage{array}
\usepackage{multirow}
\usepackage{longtable}
\usepackage{rotating}
\usepackage[table]{xcolor}
\usepackage{colortbl}
\usepackage{threeparttable} 
\usepackage{subfigure}
\usetikzlibrary{arrows.meta}
\usepackage{booktabs} 

\usetikzlibrary{shapes.geometric, arrows, arrows.meta, positioning, calc}

\definecolor{PUBlue}{RGB}{70,130,180}
\definecolor{PUGreen}{RGB}{70,160,120}
\definecolor{PUGray}{RGB}{235,235,235}
\definecolor{PUOrange}{RGB}{230,145,56}

\tikzset{
    methodfig/.style={x=0.82cm, y=0.82cm, >=stealth, font=\scriptsize},
    methodblock/.style={rounded corners=2pt, draw=black!70, line width=0.7pt,
        minimum width=1.95cm, minimum height=0.45cm, align=center, font=\scriptsize},
    convblock/.style={methodblock, fill=PUBlue!10},
    publock/.style={methodblock, fill=PUGreen!12},
    skipblock/.style={methodblock, fill=gray!12, minimum width=1.35cm},
    opblock/.style={methodblock, fill=PUGray},
    hostbox/.style={rounded corners=2pt, draw=black!70, line width=0.7pt,
        minimum width=3.00cm, minimum height=2.2cm, align=center, font=\scriptsize},
    miniblock/.style={rounded corners=2pt, draw=black!70, line width=0.7pt,
        minimum width=1.95cm, minimum height=0.45cm, align=center, font=\scriptsize},
    cellbox/.style={draw=black!70, line width=0.6pt, minimum width=0.54cm,
        minimum height=0.54cm, inner sep=0pt, font=\scriptsize},
    sumdot/.style={circle, draw=black!70, line width=0.7pt, minimum size=0.38cm,
        inner sep=0pt, font=\scriptsize},
    methodarrow/.style={->, line width=0.9pt, draw=black!85},
    skiparrow/.style={methodarrow, rounded corners=4pt}
}

\newif\ifreview
\reviewfalse

\ifreview
	\usepackage{lineno}

	\linenumbers
\fi

\begin{document}


\def\SubNumber{8}

\def\GCPRTrack{Main Track}

\title{PURe: A Plug-and-Play Product-Unit Residual Module for Vision Networks}

\ifreview
\else

	\author{Ziyuan Li\inst{1,2}\thanks{Corresponding Author} \and
    Uwe Jaekel\inst{1} \and
    Babette Dellen\inst{1}}

    \authorrunning{Li et al.}

    \institute{
    Department of Mathematics, Informatics and Technology,
    University of Applied Sciences Koblenz,
    Joseph-Rovan-Allee 2, 53424 Remagen, Germany\\
    \email{jaekel@hs-koblenz.de, dellen@hs-koblenz.de}
    \and
    Technical University of Munich, Munich, Germany\\
    \email{ziyuan.li@tum.de}
}
\fi

\maketitle              

\begin{abstract}
Modern vision networks are dominated by additive local transformations, whereas explicit multiplicative local interactions remain underexplored. Product units offer a direct approach to modeling such interactions, but their use in deep architectures has been limited by optimization instability. In this work, we propose PURe, a product-unit Residual Module for deep vision networks. PURe is built around a 2D product unit with a real-valued log-domain formulation that makes multiplicative local aggregation practical within deep residual hierarchies. The resulting module serves as a drop-in replacement for native residual units. We instantiate PURe in residual CNNs for image classification and in 2D residual encoder--decoder networks for slice-based segmentation on volumetric CT data. Across Galaxy10 DECaLS, ImageNet, and CIFAR-10, PURe consistently improves residual CNNs and yields a more favorable accuracy--parameter trade-off, allowing moderately deep models to match or surpass substantially deeper ResNet baselines with much smaller parameter budgets. On the AMOS benchmark, PURe also improves slice-based CT segmentation under 3D case-level evaluation. These results show that explicit multiplicative local interaction is a practical and effective design primitive for deep residual vision networks.

\end{abstract}
\section{Introduction}
Modern vision networks are highly expressive, yet their local feature transformations are still dominated by additive operations. Convolution aggregates spatial neighborhoods through weighted sums, and residual blocks combine features through additive shortcuts \cite{lecun2015deep,he2016deep,he2016identity,liu2022convnet}. This design has been remarkably successful, but it also imposes a strong inductive bias: higher-order local interactions are represented only indirectly through repeated compositions of linear combinations and pointwise nonlinearities. As a result, multiplicative, ratio-like, and power-law couplings are rarely modeled explicitly in deep vision architectures, despite their natural relevance to structured visual patterns.

Product units offer a straightforward approach to modeling such interactions. By coupling inputs multiplicatively with learnable exponents, they can represent structured nonlinear dependencies that are cumbersome for purely additive units \cite{durbin1989product,leerink1994learning,dellen2019function}. Recent work has demonstrated their potential in predicting physical quantities and geometric modeling \cite{dellen2024predicting,li2024data}. However, product units have not yet become common building blocks for modern deep vision systems. Their main limitation is optimization: multiplicative transformations are highly sensitive to input scale, can induce extreme gradients, and are notoriously unstable in deep architectures \cite{janson1993training,godfrey2018leveraging,engelbrecht2024fitness}. Consequently, existing product-unit applications remain mostly shallow, low-dimensional, or task-specific.

To make product units useful in modern vision networks, they must be reformulated for local feature grids, adapted to real-valued intermediate representations, and integrated into residual designs that remain usable in deep feature hierarchies. We address these requirements with PURe, a \textbf{p}roduct-\textbf{u}nit \textbf{re}sidual module for deep vision networks. PURe is built around a 2D product unit with a real-valued log-domain implementation that makes the operator practical in deep residual architectures. We then embed this operator into a residual module while preserving the shortcut pathway of the host network. The resulting design is generic and can be instantiated in both residual CNNs and 2D residual encoder--decoder networks.

Our contributions are threefold: \textbf{(i)} We generalize product units to local 2D feature grids and derive a real-valued log-domain implementation for deep feature hierarchies; \textbf{(ii)} we propose PURe, a plug-and-play residual module which replaces native spatial residual transformations with an explicit multiplicative alternative; and \textbf{(iii)} we show empirically that PURe improves residual CNNs not only in depth-matched comparisons, but also in accuracy--parameter trade-off, allowing comparatively shallow or moderately deep models to match or surpass much deeper ResNet baselines, while also transferring effectively to slice-based CT segmentation in residual encoder--decoder networks.

\section{Related Work}

Residual learning is a central design principle in modern deep networks. ResNet and its pre-activation variants showed that identity shortcuts substantially facilitate the optimization of deep models, and residual connections have since become a core component of modern ConvNet design \cite{he2016deep,he2016identity,liu2022convnet}. Our work follows this paradigm while modifying the local transformation within the residual branch. Complementary ConvNet modules such as SENet, ResNeXt, and MixConv improve feature transformations through channel recalibration, increased transformation cardinality, or mixed receptive fields, respectively \cite{hu2018squeeze,xie2017aggregated,tan2019mixconv}.

Product units were originally introduced as multiplicative counterparts to summation-based neurons \cite{durbin1989product,leerink1994learning}. They can directly represent products, powers, roots, and other structured nonlinear couplings \cite{dellen2019function}, and have recently shown promise in scientific regression and geometric modeling \cite{dellen2024predicting,li2024data}. However, their use in deep architectures remains limited because multiplicative transformations are difficult to optimize \cite{janson1993training,engelbrecht2024fitness}. Godfrey and Gashler proposed windowed product-unit neural networks, in which unweighted products over small input windows serve as nonlinear activations between conventional additive layers \cite{godfrey2018leveraging}. PURe instead uses learnable exponents and places the resulting product unit as a learnable spatial transformation within a residual branch rather than as an unweighted product activation.

Recent deep architectures have revisited explicit multiplicative interactions through polynomial and higher-order constructions. The $\Pi$-Net framework and its journal extension construct finite-order polynomial mappings using tensor-factorized parameterizations and specialized skip connections \cite{chrysos2020pinets,chrysos2022deep}. Higher-order convolution extends the conventional convolution operator through a Volterra-like expansion containing learned interaction terms of different integer orders \cite{azeglio2025convolution}. These approaches construct finite-order polynomial expansions by combining multiple multiplicative terms additively. In contrast, the 2D product unit in PURe realizes a generalized monomial transformation over a local feature neighborhood, using learnable real-valued exponents and a log-domain implementation for real-valued feature maps.

Multiplicative interactions also provide a unifying perspective on gating, attention, hypernetworks, and dynamic convolutions, particularly when multiple information or computation streams are combined \cite{jayakumar2020multiplicative}. Unlike common attention mechanisms, PURe does not construct input-dependent pairwise affinities. Instead, its exponent kernels are learned model parameters shared across spatial locations and directly aggregate local feature neighborhoods. Consequently, PURe is neither an unweighted product activation, a finite-order polynomial network, nor an attention mechanism. It is an architectural drop-in replacement for native spatial transformations within residual units, combining explicit multiplicative local aggregation with an otherwise unchanged residual composition.

\section{Method}
\subsection{2D Product Unit}
We extend the classical product unit from vector-valued inputs to local 2D feature grids. For a kernel of size $(2r+1)\times(2r+1)$, the proposed 2D product unit performs multiplicative local aggregation:
\begin{equation}
 y(i,j)=\prod_{m=-r}^{r}\prod_{n=-r}^{r} x(i+m,j+n)^{w(m,n)}.
\label{eq:2dpu}
\end{equation}
This is the multiplicative analog of local convolution, replacing the weighted summation with learnable exponentiated products in the local neighborhood.

For efficient implementation, Eq.~\eqref{eq:2dpu} can be rewritten in the log domain as
\begin{equation}
 y(i,j)=\exp\!\left(\sum_{m=-r}^{r}\sum_{n=-r}^{r} w(m,n)\,\log\!\left(x(i+m,j+n)\right)\right).
\label{eq:2dpu_logexp}
\end{equation}
For real-valued feature maps, however, negative inputs would yield complex logarithms, and extremely small positive values could produce unstable responses. We therefore introduce a trainable threshold $\tau_\theta = \mathrm{softplus}(\theta) + 10^{-7}$, and clamp the inputs before taking the logarithm:
\begin{equation}
 y(i,j)=\exp\!\left(\sum_{m=-r}^{r}\sum_{n=-r}^{r} w(m,n)\,\log\!\big(\max(x(i+m,j+n),\tau_\theta)\big)\right).
\label{eq:stable_2dpu}
\end{equation}
This keeps the computation in the real-valued domain while reducing the destabilizing effect of extremely small inputs. Since $\tau_\theta$ is learned rather than fixed, the operator can adapt to the scale of intermediate feature maps during training.

\begin{figure}[h]
    \centering
    \includegraphics[width=0.43\linewidth]{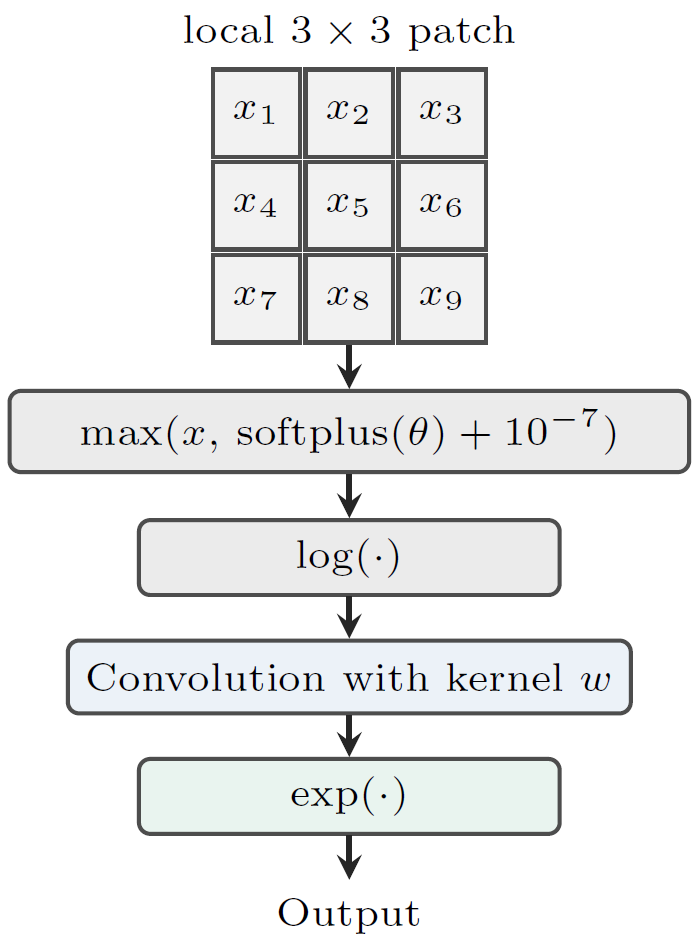}
    \caption{Computational flow of the proposed 2D product unit for real-valued feature maps. Inputs are lower-bounded by a learnable threshold, transformed to the log domain, aggregated by convolution, and mapped back by exponentiation.}
    \label{fig:2dpu_flow}
\end{figure}

In practice, the 2D product unit performs the following computations: \textbf{(i)} apply $\log(\max(x,\tau_\theta))$; \textbf{(ii)} aggregate in log space with a standard convolution kernel; and \textbf{(iii)} map the response back by exponentiation. The computational flow is shown in Fig.~\ref{fig:2dpu_flow}.

\subsection{PURe: Product-Unit Residual Module}
We next embed the 2D product unit into a residual design. The key idea is to keep the shortcut pathway unchanged while replacing one native spatial transformation in the residual branch with a 2D product unit. In addition, the intermediate ReLU activations inside the original residual branch are removed, since interleaved pointwise nonlinearities would interfere with the intended multiplicative transformation. The shortcut therefore retains the optimization benefits of residual learning, whereas the transformed branch introduces explicit multiplicative local interaction.

For an input feature tensor $x$, a generic PURe can be written as
\begin{equation}
 y = \mathcal{S}(x) + \mathcal{F}_{\mathrm{PU}}(x),
\label{eq:pure_generic}
\end{equation}
where $\mathcal{F}_{\mathrm{PU}}(\cdot)$ denotes a host-dependent residual branch containing a 2D product unit, and $\mathcal{S}(\cdot)$ denotes the shortcut mapping. In the simplest case, $\mathcal{S}(x)=x$; when the input and output dimensions differ, $\mathcal{S}(\cdot)$ becomes a projection branch, exactly as in standard residual designs.

\begin{figure}[h]
    \centering
    \includegraphics[width=0.89\linewidth]{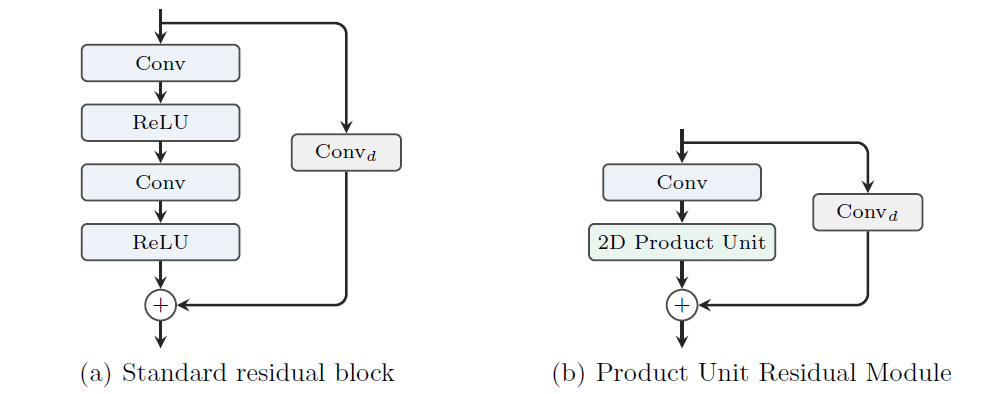}
    \caption{Standard residual block versus PURe. PURe keeps the shortcut mapping unchanged, replaces the second spatial transformation in the residual branch with a 2D product unit, and removes the intermediate ReLU activations inside the branch.}
    \label{fig:pure_block}
\end{figure}

PURe therefore changes the local transformation inside the residual unit while leaving the outer residual composition unchanged. A schematic comparison between a standard residual block and PURe is shown in Fig.~\ref{fig:pure_block}. This makes the proposed product-unit operator usable as a plug-and-play residual module across different host architectures.

\subsection{Instantiations in ResNet and Residual U-Net}
In ResNet-style classifiers, PURe is introduced at the level of the native residual block. For the basic-block configuration used in ResNet18/34, the first $3\times3$ convolution is kept unchanged, and the second $3\times3$ convolution in the residual branch is replaced by a 2D product unit. For the bottleneck configuration used in ResNet50/101/152, the two surrounding $1\times1$ convolutions are preserved, and only the central spatial-mixing layer is replaced by a 2D product unit. In both cases, the shortcut branch, including projection shortcuts for channel expansion or spatial downsampling, is kept identical to that of the corresponding baseline, while the ReLU activations inside the original residual branch are removed.

For slice-based segmentation on volumetric CT data, we apply the same replacement principle to the native residual units in the encoder, bottleneck, and decoder of a residual U-Net. The overall ResUNet macro-architecture is otherwise kept fixed, including the resolution transitions, long skip connections, channel configuration, and prediction head. As a result, the comparison between ResUNet and PURe-ResUNet differs only in the formulation of the local residual unit, rather than in the surrounding multi-scale encoder--decoder structure.

\section{Experiments}
We evaluate PURe in two complementary settings to assess both its practical effectiveness and its architectural scope. The first considers image classification in deep residual CNNs, and the second considers image segmentation in residual encoder--decoder networks.

\subsection{Classification on residual CNNs}
We first evaluate PURe on three classification benchmarks with different characteristics: Galaxy10 DECaLS, a medium-scale galaxy morphology dataset\cite{willett2013galaxyzoo2,dey2019legacysurveys}; ImageNet, a large-scale natural-image classification benchmark \cite{deng2009imagenet}; and CIFAR-10 \cite{krizhevsky2009learning}, a standard low-resolution classification dataset.

\subsubsection{Galaxy10 DECaLS}
\paragraph{\textbf{Dataset.}}
Galaxy10 DECaLS \cite{willett2013galaxyzoo2,dey2019legacysurveys} is derived from the Dark Energy Camera Legacy Survey and contains approximately 17,736 RGB images from 10 galaxy classes, each of size $256\times256$. The dataset covers diverse galaxy morphologies, including disturbed, merging, barred spiral, and unbarred spiral types, and therefore provides a structured classification setting with substantial intra-class and inter-class variation.

\paragraph{\textbf{Implementation.}}
We use a stratified train/val/test split of $0.90{:}0.05{:}0.05$. Data augmentation is applied only to the training set and includes random rotation ($\pm20^\circ$), horizontal flipping, and random resized cropping.

All models are trained with SGD (batch size 64, momentum 0.9) for 30 epochs using cross-entropy loss. Because standard ResNet blocks and PURe blocks exhibit different optimization behavior, we tune the training strategy separately for the two architecture families. The ResNet baselines use an initial learning rate of 0.1 and weight decay of 0.001, whereas the PURe-based models use an initial learning rate of 0.01 and weight decay of 0.01, consistent with prior observations that product-unit networks benefit from smaller learning rates \cite{dellen2019function}. For both families, the learning rate is reduced by a factor of 0.1 when the validation loss does not improve for three consecutive epochs, and the checkpoint with the lowest validation loss is used for final evaluation.

Each network is trained five times. Robustness is additionally evaluated on a Poisson-corrupted test set that mimics photon noise in astronomical imaging \cite{vio2005least}, and the relative performance drop is measured with respect to the original test-set accuracy.

\paragraph{\textbf{Results and Analysis.}}
Table~\ref{tab:galaxy10_results} shows that PURe consistently improves residual CNNs on Galaxy10 DECaLS. PURe-ResNet34 achieves the strongest clean-set performance, outperforming all ResNet baselines while using substantially fewer parameters than much deeper models such as ResNet152. At the smaller model scale, PURe-ResNet18 also improves over its ResNet counterpart. These results indicate that PURe improves not only depth-matched performance, but also the accuracy--parameter trade-off.

\begin{table}[h]
\caption{Classification results on Galaxy10 DECaLS. Original-set accuracy is reported as mean$\pm$std over five runs; all remaining metrics are averaged across runs. Params denotes the number of parameters. Clean and Noisy denote accuracy on the original and Poisson-corrupted test sets, respectively. Drop denotes the relative performance drop from Clean to Noisy. T80 denotes the training time required to first reach 80\% validation accuracy.}
\centering
\small
\setlength{\tabcolsep}{2pt}
\begin{tabular}{lcccccc}
\hline
Model & Params (M) & Clean (\%) & Noisy (\%) & Drop (\%) & T80 (s) & Train (s) \\
\hline
PURe-ResNet18 & 11.18 & 83.63 $\pm$ 0.65          & 83.04          & 0.70          & \textbf{444.22} & 1211.51          \\
PURe-ResNet34 & 21.29 & \textbf{84.28 $\pm$ 0.56} & \textbf{83.83} & \textbf{0.54} & 541.93          & 1593.92          \\
\hline
ResNet18      & 11.18 & 82.07 $\pm$ 1.28          & 81.38          & 0.85          & 528.50          & \textbf{1148.91} \\
ResNet34      & 21.29 & 82.84 $\pm$ 0.65          & 82.01          & 1.01          & 635.38          & 1537.20          \\
ResNet50      & 23.53 & 83.68 $\pm$ 0.89          & 82.86          & 0.97          & 1523.09         & 2456.60          \\
ResNet101     & 42.52 & 83.20 $\pm$ 0.79          & 82.59          & 0.73          & 2330.46         & 3884.10          \\
ResNet152     & 58.16 & 83.86 $\pm$ 0.61          & 83.16          & 0.83          & 3817.27         & 5351.32          \\
\hline
\end{tabular}
\label{tab:galaxy10_results}
\end{table}

PURe-based models are also more robust under Poisson corruption. PURe-ResNet34 attains the highest noisy-set accuracy, and PURe-ResNet18 reaches the 80\% validation-accuracy threshold faster than all ResNet baselines. Overall, the Galaxy10 results show that PURe improves residual CNNs in accuracy, parameter efficiency, robustness, and optimization speed.

For context, Table~\ref{tab:galaxy10_literature} lists previously reported ResNet accuracies on Galaxy10 DECaLS under different augmentation pipelines. They provide a reference range for conventional ResNet performance. Our ResNet baselines are broadly consistent with prior reports, while PURe-ResNet34 achieves a clearly higher accuracy.

\begin{table}[ht]
\caption{Previously reported ResNet accuracies on Galaxy10 DECaLS under different augmentation pipelines.}
\setlength{\tabcolsep}{5pt}
\centering
\footnotesize
\begin{tabular}{lcc}
\hline
Augmentation setting & Model & Acc. (\%) \\
\hline
FSL-GAN transformations \cite{yao2024galaxy} 
& ResNet18 & 76.81 \\
\hline
\multirow{4}{*}{Randomized single-image transformations \cite{le2025enhanced}}
& ResNet50  & 78.88 \\
& ResNet101 & 79.99 \\
& ResNet152 & 80.61 \\
\midrule
Basic single-image transformations \cite{cheng2023application}
& ResNet50 & 80.00 \\
\hline
\end{tabular}
\label{tab:galaxy10_literature}
\end{table}

\subsubsection{ImageNet}
\paragraph{\textbf{Dataset and Implementation.}}
We evaluate PURe on ImageNet \cite{deng2009imagenet}, which contains approximately 1.28 million training images and 50,000 validation images from 1,000 classes. We consider PURe-ResNet18, PURe-ResNet34, and PURe-ResNet50, obtained by replacing the native residual blocks in the corresponding ResNet architectures with PURe blocks while preserving the original stage hierarchy and shortcut design. Data preprocessing and augmentation follow standard ResNet practice \cite{he2016deep}. All models are trained with SGD and momentum 0.9 for 90 epochs, using an initial learning rate of 0.01 and weight decay of 0.001, with learning-rate decay at epochs 30 and 60.

\paragraph{\textbf{Results and Analysis.}}
\begin{table}[h]
\caption{ImageNet-1K validation accuracy of PURe-based ResNets and standard ResNet baselines. ResNet baseline numbers are taken from He et al.~\cite{he2016deep}.}
\setlength{\tabcolsep}{5pt}
\centering
\small
\begin{tabular}{@{}lccc@{}}
\hline
Model & Top-1 Acc. (\%) & Top-5 Acc. (\%) & \#Params (M) \\
\hline
PURe-ResNet18 & 78.21 & 95.04 & 11.69 \\
PURe-ResNet34 & 80.27 & 95.78 & 21.80 \\
PURe-ResNet50 & \textbf{81.23} & \textbf{96.18} & 25.56 \\
\hline
ResNet34 \cite{he2016deep}  & 78.16 & 94.29 & 21.80 \\
ResNet50 \cite{he2016deep}  & 79.26 & 94.75 & 25.56 \\
ResNet101 \cite{he2016deep} & 80.13 & 95.40 & 44.55 \\
ResNet152 \cite{he2016deep} & 80.62 & 95.51 & 60.19 \\
\hline
\end{tabular}
\label{tab:imagenet_results}
\end{table}

Table~\ref{tab:imagenet_results} shows that PURe consistently improves residual CNNs on ImageNet. PURe-ResNet18 exceeds the reported ResNet34 baseline with roughly half the parameters, while PURe-ResNet34 surpasses the reported ResNet50 and ResNet101 baselines and remains close to ResNet152 with a substantially smaller parameter budget. Most notably, PURe-ResNet50 achieves the strongest overall performance, exceeding the much deeper ResNet152 baseline while using less than half the parameters. These results show that PURe improves not only depth-matched residual CNNs, but also the accuracy--parameter trade-off on large-scale natural-image classification.

\subsubsection{CIFAR-10}
\paragraph{\textbf{Dataset and Implementation.}}
We further evaluate PURe on CIFAR-10 \cite{krizhevsky2009learning}, which contains 60,000 color images of size $32\times32$ from 10 classes. Following the standard CIFAR-style ResNet design \cite{he2016deep}, we replace the ImageNet-style stem with a single $3\times3$ convolution, omit the fourth residual stage, and consider PURe-ResNet20, PURe-ResNet32, PURe-ResNet44, PURe-ResNet56, PURe-ResNet110, and PURe-ResNet272.

All PURe-based models are trained with SGD (batch size 128, momentum 0.9) and cross-entropy loss. The standard schedule uses 160 epochs with an initial learning rate of 0.01, weight decay of 0.001, and learning-rate decay at epochs 80 and 120. For PURe-ResNet110 and PURe-ResNet272, we also consider an extended 220-epoch schedule with learning-rate decay at epochs 140 and 180. Each model is trained five times and the results are reported as mean $\pm$ standard deviation. ResNet references are from He et al.~\cite{he2016deep,he2016identity}.

\paragraph{\textbf{Results and Analysis.}}
Under the standard 160-epoch schedule, Table~\ref{tab:cifar_results} shows that PURe-based residual CNNs consistently outperform CIFAR-style ResNet counterparts across all evaluated depths. The gains are already visible in shallow models and remain present at greater depth, indicating that replacing additive residual blocks with PURe blocks yields stable improvements throughout the standard CIFAR-style residual CNN regime.

\begin{table}[h]
\caption{CIFAR-10 classification accuracy of PURe-based ResNets. Standard and deeper pre-activation ResNet references are from He et al.~\cite{he2016deep,he2016identity}.}
\centering
\small
\setlength{\tabcolsep}{4pt}
\begin{tabular}{lcccc}
\toprule
\multicolumn{2}{c}{PURe-based model} & \multicolumn{2}{c}{ResNet reference} & \#Params (M) \\
\cmidrule(lr){1-2}\cmidrule(lr){3-4}
Model & Acc. (\%) & Model & Acc. (\%) &  \\
\midrule
\multicolumn{5}{c}{\textit{Standard 160-epoch schedule}} \\
\midrule
PURe-ResNet20  & \textbf{91.80 $\pm$ 0.17} & ResNet20 \cite{he2016deep}  & 91.25            & 0.27 \\
PURe-ResNet32  & \textbf{92.83 $\pm$ 0.18} & ResNet32 \cite{he2016deep}  & 92.49            & 0.46 \\
PURe-ResNet44  & \textbf{93.21 $\pm$ 0.16} & ResNet44 \cite{he2016deep}  & 92.83            & 0.66 \\
PURe-ResNet56  & \textbf{93.51 $\pm$ 0.12} & ResNet56 \cite{he2016deep}  & 93.03            & 0.85 \\
PURe-ResNet110 & \textbf{93.93 $\pm$ 0.09} & ResNet110 \cite{he2016deep} & 93.39 $\pm$ 0.16 & 1.73 \\
\midrule
\multicolumn{5}{c}{\textit{Extended-schedule and deeper reference results}} \\
\midrule
PURe-ResNet110 & \textbf{94.61 $\pm$ 0.11} & ResNet110 \cite{he2016identity}  & 93.63              & 1.73 \\
\multicolumn{2}{c}{--} & ResNet164 \cite{he2016identity}  & 94.54              & 1.76 \\
PURe-ResNet272 & 95.01 $\pm$ 0.09          & \multicolumn{2}{c}{--}          & 4.36 \\
\multicolumn{2}{c}{--} & ResNet1001 \cite{he2016identity} & 95.11 $\pm$ 0.14   & 10.15 \\
\bottomrule
\end{tabular}
\label{tab:cifar_results}
\end{table}

Table~\ref{tab:cifar_results} also places these results in a broader depth-scaling and parameter-efficiency context by including extended-schedule results for deeper PURe-based models together with deeper pre-activation ResNet references \cite{he2016identity}. Most notably, PURe-ResNet272 approaches the much deeper ResNet1001 baseline while using less than half the parameters. Thus, the advantage of PURe on CIFAR-10 is not limited to depth-matched gains, but also extends to a more favorable accuracy--parameter trade-off at greater depth.

\subsection{Segmentation on residual encoder--decoder networks}
We next evaluate PURe on image segmentation to examine whether the proposed residual module remains effective for dense prediction in residual encoder--decoder architectures.

\paragraph{\textbf{Dataset.}}
For segmentation, we use the CT subset of the AMOS benchmark \cite{ji2022amos}. AMOS is a large-scale abdominal multi-organ segmentation benchmark with voxel-level annotations for 15 anatomical structures, including abdominal organs and major vessels. The CT subset exhibits substantial variability in anatomy, contrast, field of view, and surrounding tissue context, making it a challenging benchmark for dense prediction. We restrict our study to the CT portion in order to maintain modality consistency and isolate the effect of PURe on residual encoder--decoder networks.

For each CT volume, we extract 2D slices and their corresponding label maps from the three canonical anatomical directions: axial, coronal, and sagittal. These slices form the training and evaluation samples for the 2D residual encoder--decoder networks.

\paragraph{\textbf{Implementation.}}
We use the official AMOS CT training set for model development, hold out 5\% for validation, and use the official AMOS CT validation set as the final test set.

Both models are trained under the same slice-based protocol. For each CT volume, 2D image--label pairs are extracted along the axial, coronal, and sagittal directions, pooled into a single training set, and used to train a shared 2D residual encoder--decoder network. Training uses AdamW, a batch size of 2 per GPU, cosine learning-rate decay over 300 epochs, mixed precision, and gradient clipping at 12.0. The ResUNet baseline uses a learning rate of $2\times10^{-4}$ and weight decay of $10^{-4}$, while PURe-ResUNet uses the same setting except for a 10$\times$ smaller learning rate. We further apply random flipping, in-plane rotation, Gaussian noise, and intensity scaling and shifting, and optimize a combined Dice and cross-entropy loss with equal weights.

At inference time, each test volume is processed slice by slice along the three directions, and the resulting 2D predictions are reassembled into 3D probability volumes in the original case space. Final predictions are obtained by averaging class probabilities across views followed by voxel-wise label assignment.

Performance is evaluated on a case-wise basis. For each test case, we compute Dice scores for all 15 anatomical classes on the reconstructed 3D segmentation and define mDice as the average across classes or, when necessary, across non-empty classes only. We report mean case-wise mDice with 95\% confidence intervals. Statistical significance is assessed with paired Wilcoxon signed-rank tests \cite{wilcoxon1992individual}, with Benjamini--Hochberg correction applied for class-wise comparisons \cite{benjamini1995controlling}.

\paragraph{\textbf{Results and Analysis.}}
Table~\ref{tab:amos_ct_results} shows that PURe-ResUNet consistently outperforms the baseline ResUNet on the AMOS CT validation set. At the overall level, the gain in case-wise mDice is statistically significant under paired testing, indicating that the improvement is systematic across cases rather than driven by a small number of outliers.

\begin{table}[h]
\setlength{\tabcolsep}{3pt}
\centering
\caption{Paired 3D evaluation on the AMOS CT validation set. The first row reports overall mDice, whereas the remaining rows report class-wise Dice. Values are shown as mean [95\% CI]. $\Delta$ denotes PURe-ResUNet minus ResUNet in percentage points (pp). Class-wise $p$-values are adjusted using the Benjamini--Hochberg FDR procedure.}
\label{tab:amos_ct_results}
\resizebox{\textwidth}{!}{%
\begin{tabular}{lccccc}
\toprule
Structure & ResUNet Dice (\%) & PURe-ResUNet (\%) & $\Delta$ (pp) & Wilcoxon $p$ & FDR $p$ \\
\midrule
Overall mDice & 82.3 [81.1, 83.5] & 83.7 [82.6, 84.9] & +1.5 [1.1, 1.8] & $1.1\times10^{-13}$ & -- \\
Spleen & 93.8 [92.3, 95.2] & 95.1 [93.9, 96.2] & +1.3 [0.6, 2.0] & $3.2\times10^{-14}$ & $2.4\times10^{-13}$ \\
Right kidney & 92.0 [89.2, 94.7] & 94.0 [91.8, 96.1] & +2.0 [0.4, 3.5] & $5.3\times10^{-13}$ & $2.6\times10^{-12}$ \\
Left kidney & 92.7 [91.3, 94.2] & 94.5 [93.1, 96.0] & +1.8 [1.0, 2.6] & $9.0\times10^{-13}$ & $3.4\times10^{-12}$ \\
Gallbladder & 76.7 [71.7, 81.7] & 77.2 [71.9, 82.6] & +0.5 [-1.1, 2.2] & $1.7\times10^{-5}$ & $2.3\times10^{-5}$ \\
Esophagus & 78.1 [75.9, 80.3] & 78.8 [76.2, 81.3] & +0.7 [-0.2, 1.6] & $5.0\times10^{-4}$ & $6.0\times10^{-4}$ \\
Liver & 95.7 [95.0, 96.4] & 96.8 [96.1, 97.5] & +1.1 [0.8, 1.3] & $8.0\times10^{-17}$ & $1.2\times10^{-15}$ \\
Stomach & 85.6 [82.4, 88.8] & 87.1 [83.9, 90.2] & +1.5 [0.6, 2.3] & $1.6\times10^{-6}$ & $3.0\times10^{-6}$ \\
Aorta & 93.0 [92.0, 94.0] & 93.7 [92.6, 94.9] & +0.8 [0.4, 1.1] & $9.2\times10^{-10}$ & $2.7\times10^{-9}$ \\
Inferior vena cava & 86.1 [85.0, 87.3] & 87.4 [86.2, 88.6] & +1.3 [0.8, 1.7] & $7.8\times10^{-9}$ & $1.7\times10^{-8}$ \\
Pancreas & 79.4 [77.0, 81.7] & 81.2 [78.8, 83.6] & +1.8 [1.1, 2.5] & $7.8\times10^{-9}$ & $1.7\times10^{-8}$ \\
Right adrenal gland & 69.8 [67.3, 72.3] & 71.8 [69.5, 74.1] & +2.0 [0.8, 3.2] & $2.2\times10^{-5}$ & $2.7\times10^{-5}$ \\
Left adrenal gland & 66.3 [63.0, 69.5] & 70.0 [67.1, 72.9] & +3.8 [2.1, 5.4] & $4.8\times10^{-6}$ & $8.0\times10^{-6}$ \\
Duodenum & 71.2 [68.1, 74.4] & 73.0 [70.0, 76.1] & +1.8 [1.1, 2.6] & $6.5\times10^{-6}$ & $9.7\times10^{-6}$ \\
Bladder & 78.6 [74.0, 83.2] & 79.4 [74.8, 84.0] & +0.8 [-0.7, 2.2] & 0.0046 & 0.0050 \\
Prostate/Uterus & 75.6 [71.7, 79.4] & 76.2 [72.5, 80.0] & +0.7 [-0.9, 2.3] & 0.0159 & 0.0159 \\
\bottomrule
\end{tabular}%
}
\end{table}

At the class level, PURe-ResUNet improves all 15 anatomical structures. The gains are especially pronounced for smaller or more challenging targets, while clear improvements are also observed for larger organs, indicating that the benefit of PURe is not restricted to a particular size regime. After Benjamini--Hochberg correction, all class-wise comparisons remain statistically significant, suggesting that the advantage of PURe is broad and anatomically consistent.

\begin{figure}
    \centering
    \includegraphics[width=0.94\linewidth]{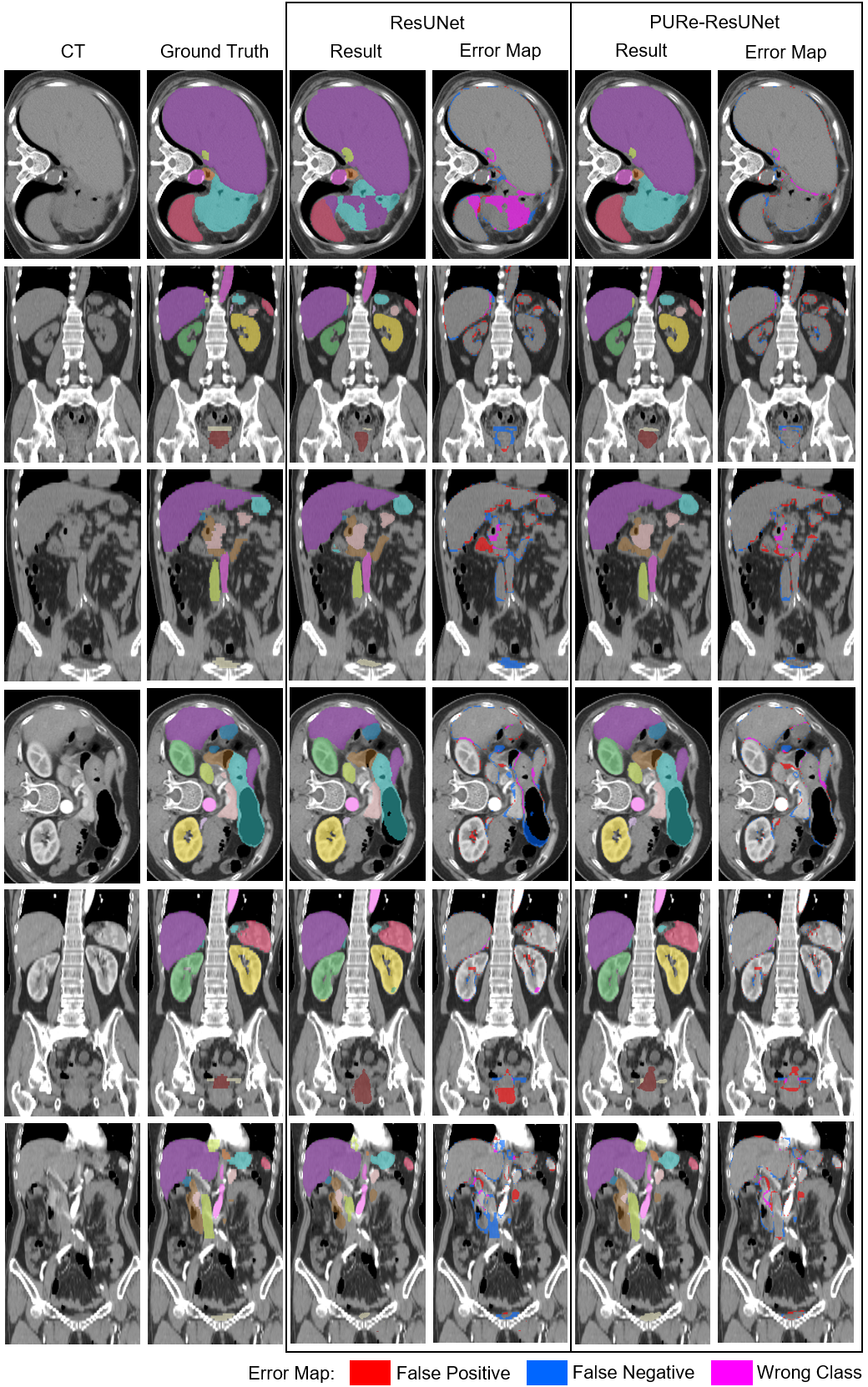}
    \caption{Qualitative comparison on the AMOS CT validation set.}
    \label{fig:amos_ct_qualitative}
\end{figure}

Figure~\ref{fig:amos_ct_qualitative} provides representative qualitative comparisons on the AMOS CT validation set. Relative to the baseline ResUNet, PURe-ResUNet typically yields more complete organ regions, cleaner boundaries, and fewer wrong-class predictions after reconstructing the 3D case-level segmentation from tri-planar slice predictions. The reduction in error is particularly visible in anatomically small or visually ambiguous structures, where local context and boundary precision are especially important.

Overall, these results show that the proposed multiplicative residual formulation transfers effectively from classification backbones to slice-based CT segmentation and is particularly helpful for structures with complex boundaries, limited spatial extent, or strong contextual ambiguity.

\section{Discussion and Conclusion}
Across both classification and segmentation, PURe consistently improves residual vision networks without changing their surrounding macro-architecture. In residual CNNs, the gains appear not only in depth-matched comparisons, but also in a more favorable accuracy--parameter trade-off, with PURe-based models often approaching or surpassing substantially deeper ResNet baselines using fewer parameters. On AMOS CT, the same residual formulation improves overall case-wise Dice under 3D case-level evaluation and yields gains across all anatomical classes, indicating that the benefit of PURe extends beyond image classification.

A natural interpretation is that PURe strengthens the residual branch by introducing explicit local multiplicative interactions. In conventional residual units, such interactions must be represented indirectly through repeated compositions of additive operators and pointwise nonlinearities. PURe instead incorporates them directly into the residual transformation, while the shortcut pathway preserves the optimization advantages of residual learning and makes the multiplicative branch practical in deep architectures.

Overall, these results suggest that explicit multiplicative local interactions offer a viable and effective design primitive for residual vision models. Rather than introducing a new backbone, PURe provides a simple plug-and-play way to enrich local feature transformation within established residual architectures, offering a practical route toward stronger and more parameter-efficient deep vision networks.

%
%
%
%
\bibliographystyle{splncs04}
\bibliography{reference}

@inproceedings{he2016deep,
  title={Deep residual learning for image recognition},
  author={He, Kaiming and Zhang, Xiangyu and Ren, Shaoqing and Sun, Jian},
  booktitle={Proceedings of the IEEE conference on computer vision and pattern recognition},
  pages={770--778},
  year={2016}
}

@inproceedings{he2016identity,
  title={Identity mappings in deep residual networks},
  author={He, Kaiming and Zhang, Xiangyu and Ren, Shaoqing and Sun, Jian},
  booktitle={Computer Vision--ECCV 2016: 14th European Conference, Amsterdam, The Netherlands, October 11--14, 2016, Proceedings, Part IV 14},
  pages={630--645},
  year={2016},
  organization={Springer}
}

@inproceedings{hu2018squeeze,
  title={Squeeze-and-excitation networks},
  author={Hu, Jie and Shen, Li and Sun, Gang},
  booktitle={Proceedings of the IEEE conference on computer vision and pattern recognition},
  pages={7132--7141},
  year={2018}
}

@inproceedings{xie2017aggregated,
  title={Aggregated residual transformations for deep neural networks},
  author={Xie, Saining and Girshick, Ross and Doll{\'a}r, Piotr and Tu, Zhuowen and He, Kaiming},
  booktitle={Proceedings of the IEEE conference on computer vision and pattern recognition},
  pages={1492--1500},
  year={2017}
}

@article{tan2019mixconv,
  title={Mixconv: Mixed depthwise convolutional kernels},
  author={Tan, Mingxing and Le, Quoc V},
  journal={arXiv preprint arXiv:1907.09595},
  year={2019}
}

@inproceedings{liu2022convnet,
  title={A convnet for the 2020s},
  author={Liu, Zhuang and Mao, Hanzi and Wu, Chao-Yuan and Feichtenhofer, Christoph and Darrell, Trevor and Xie, Saining},
  booktitle={Proceedings of the IEEE/CVF conference on computer vision and pattern recognition},
  pages={11976--11986},
  year={2022}
}

@article{durbin1989product,
  title={Product units: A computationally powerful and biologically plausible extension to backpropagation networks},
  author={Durbin, Richard and Rumelhart, David E},
  journal={Neural computation},
  volume={1},
  number={1},
  pages={133--142},
  year={1989},
  publisher={MIT Press One Rogers Street, Cambridge, MA 02142-1209, USA journals-info~…}
}

@article{leerink1994learning,
  title={Learning with product units},
  author={Leerink, Laurens and Giles, C and Horne, Bill and Jabri, Marwan},
  journal={Advances in neural information processing systems},
  volume={7},
  year={1994}
}

@article{dellen2024predicting,
  title={Predicting nuclear masses with product-unit networks},
  author={Dellen, Babette and Jaekel, Uwe and Freitas, Paulo SA and Clark, John W},
  journal={Physics Letters B},
  volume={852},
  pages={138608},
  year={2024},
  publisher={Elsevier}
}

@inproceedings{dellen2019function,
  title={Function and pattern extrapolation with product-unit networks},
  author={Dellen, Babette and Jaekel, Uwe and Wolnitza, Marcell},
  booktitle={Computational Science--ICCS 2019: 19th International Conference, Faro, Portugal, June 12--14, 2019, Proceedings, Part II 19},
  pages={174--188},
  year={2019},
  organization={Springer}
}

@inproceedings{li2024data,
  title={Data-Driven 3D Shape Completion with Product Units},
  author={Li, Ziyuan and Jaekel, Uwe and Dellen, Babette},
  booktitle={International Conference on Computational Science},
  pages={302--315},
  year={2024},
  organization={Springer}
}

@techreport{krizhevsky2009learning,
  title={Learning multiple layers of features from tiny images},
  author={Krizhevsky, Alex and Hinton, Geoffrey and others},
  year={2009},
  institution={University of Toronto}
}

@inproceedings{deng2009imagenet,
  title={Imagenet: A large-scale hierarchical image database},
  author={Deng, Jia and Dong, Wei and Socher, Richard and Li, Li-Jia and Li, Kai and Fei-Fei, Li},
  booktitle={2009 IEEE conference on computer vision and pattern recognition},
  pages={248--255},
  year={2009},
  organization={Ieee}
}

@article{le2025enhanced,
  title={Enhanced Comparative Analysis of Pretrained and Custom Deep Convolutional Neural Networks for Galaxy Morphology Classification},
  author={Le, Tram and Ibrahim, Nickson and Nguyen, Thu and Noiplab, Thanyaporn and Kim, Jungyoon and Bhati, Deepshikha},
  journal={Engineering Proceedings},
  volume={89},
  number={1},
  pages={36},
  year={2025},
  publisher={MDPI}
}

@inproceedings{cheng2023application,
  title={Application and Analysis of Residual Blocks in Galaxy Classification},
  author={Cheng, Wenzheng},
  booktitle={Applied and Computational Engineering},
  volume={21},
  pages={143--152},
  year={2023},
}

@article{yao2024galaxy,
  title={A Galaxy Image Augmentation Method Based on Few-shot Learning and Generative Adversarial Networks},
  author={Yao, Yiqi and Zhang, Jinqu and Du, Ping and Dong, Shuyu},
  journal={Research in Astronomy and Astrophysics},
  volume={24},
  number={3},
  pages={035015},
  year={2024},
  publisher={IOP Publishing}
}

@article{lecun2015deep,
  title={Deep learning},
  author={LeCun, Yann and Bengio, Yoshua and Hinton, Geoffrey},
  journal={nature},
  volume={521},
  number={7553},
  pages={436--444},
  year={2015},
  publisher={Nature Publishing Group UK London}
}

@article{engelbrecht2024fitness,
  title={Fitness Landscape Analysis of Product Unit Neural Networks},
  author={Engelbrecht, Andries and Gouldie, Robert},
  journal={Algorithms},
  volume={17},
  number={6},
  pages={241},
  year={2024},
  publisher={MDPI}
}

@article{janson1993training,
  title={Training product unit neural networks with genetic algorithms},
  author={Janson, David J and Frenzel, James F},
  journal={IEEE Expert},
  volume={8},
  number={5},
  pages={26--33},
  year={1993},
  publisher={IEEE}
}

@inproceedings{godfrey2018leveraging,
  title={Leveraging Product as an Activation Function in Deep Networks},
  author={Godfrey, Luke B and Gashler, Michael S},
  booktitle={2018 IEEE International Conference on Systems, Man, and Cybernetics (SMC)},
  pages={1617--1622},
  year={2018},
  organization={IEEE}
}

@article{willett2013galaxyzoo2,
  author  = {Willett, Kyle W. and Lintott, Chris J. and Bamford, Steven P. and others},
  title   = {Galaxy Zoo 2: detailed morphological classifications for 304,122 galaxies from the Sloan Digital Sky Survey},
  journal = {Monthly Notices of the Royal Astronomical Society},
  volume  = {435},
  number  = {4},
  pages   = {2835--2860},
  year    = {2013}
}

@article{dey2019legacysurveys,
  author  = {Dey, Arjun and Schlegel, David J. and Lang, Dustin and others},
  title   = {Overview of the DESI Legacy Imaging Surveys},
  journal = {The Astronomical Journal},
  volume  = {157},
  number  = {5},
  pages   = {168},
  year    = {2019},
}

@article{ji2022amos,
  title={Amos: A large-scale abdominal multi-organ benchmark for versatile medical image segmentation},
  author={Ji, Yuanfeng and Bai, Haotian and Ge, Chongjian and Yang, Jie and Zhu, Ye and Zhang, Ruimao and Li, Zhen and Zhang, Lingyan and Ma, Wanling and Wan, Xiang and others},
  journal={Advances in neural information processing systems},
  volume={35},
  pages={36722--36732},
  year={2022}
}

@incollection{wilcoxon1992individual,
  title={Individual comparisons by ranking methods},
  author={Wilcoxon, Frank},
  booktitle={Breakthroughs in statistics: Methodology and distribution},
  pages={196--202},
  year={1992},
  publisher={Springer}
}

@article{benjamini1995controlling,
  title={Controlling the false discovery rate: a practical and powerful approach to multiple testing},
  author={Benjamini, Yoav and Hochberg, Yosef},
  journal={Journal of the Royal statistical society: series B (Methodological)},
  volume={57},
  number={1},
  pages={289--300},
  year={1995},
  publisher={Wiley Online Library}
}

@article{vio2005least,
  title={Least-squares methods with Poissonian noise: Analysis and comparison with the Richardson-Lucy algorithm},
  author={Vio, R and Bardsley, J and Wamsteker, W},
  journal={Astronomy \& Astrophysics},
  volume={436},
  number={2},
  pages={741--755},
  year={2005},
  publisher={EDP Sciences}
}

@inproceedings{jayakumar2020multiplicative,
  title={Multiplicative interactions and where to find them},
  author={Jayakumar, Siddhant M and Czarnecki, Wojciech M and Menick, Jacob and Schwarz, Jonathan and Rae, Jack and Osindero, Simon and Teh, Yee Whye and Harley, Tim and Pascanu, Razvan},
  booktitle={International conference on learning representations},
  year={2020}
}

@article{azeglio2025convolution,
  title={Convolution goes higher-order: a biologically inspired mechanism empowers image classification},
  author={Azeglio, Simone and Marre, Olivier and Neri, Peter and Ferrari, Ulisse},
  journal={Advances in Neural Information Processing Systems},
  volume={38},
  pages={11289--11322},
  year={2025}
}

@article{chrysos2022deep,
  title={Deep polynomial neural networks},
  author={Chrysos, Grigorios G and Moschoglou, Stylianos and Bouritsas, Giorgos and Deng, Jiankang and Panagakis, Yannis and Zafeiriou, Stefanos},
  journal={IEEE transactions on pattern analysis and machine intelligence},
  volume={44},
  number={8},
  pages={4021--4034},
  year={2021},
  publisher={IEEE}
}

@inproceedings{chrysos2020pinets,
  title = {{$\Pi$-Nets: Deep Polynomial Neural Networks}},
  author={Chrysos, Grigorios G and Moschoglou, Stylianos and Bouritsas, Giorgos and Panagakis, Yannis and Deng, Jiankang and Zafeiriou, Stefanos},
  booktitle={Proceedings of the IEEE/CVF Conference on Computer Vision and Pattern Recognition},
  pages={7325--7335},
  year={2020}
}

\end{document}